\pdfoutput=1

\documentclass[11pt]{article}

\usepackage[preprint]{acl}

\usepackage{times}
\usepackage{latexsym}

\usepackage[T1]{fontenc}
\usepackage[utf8]{inputenc}

\usepackage{microtype}

\usepackage{inconsolata}

\usepackage{graphicx}

\usepackage{booktabs}

\usepackage{makecell}
\usepackage{pifont}
\newcommand{\YES}{\ding{51}}
\newcommand{\NO}{{\color{gray!50}\ding{55}}}
\newenvironment{citemize}{\begin{list}{\scriptsize$\bullet$}{\topsep=.5\smallskipamount\itemsep=2pt\parsep=1pt\labelwidth=1em}}{\end{list}}

\title{NameTag 3: A Tool and a Service for Multilingual/Multitagset NER}

\author{Jana Straková \and Milan Straka\\
  Charles University, Faculty of Mathematics and Physics,\\
  Institute of Formal and Applied Linguistics \\
  Malostranské nám. 25, Prague, Czech Republic \\
  \texttt{\{strakova,straka\}@ufal.mff.cuni.cz}}

\usepackage{fancyhdr}
\fancypagestyle{officialbibref}{\fancyhf{}\fancyheadoffset[L,R]{50pt}\fancyhead[C]{This paper was published at \textbf{ACL 2025} -- please cite the published version {\small\url{https://aclanthology.org/2025.acl-demo.4/}}.}\fancyfoot[C]{\normalsize\thepage}\addtolength{\topmargin}{-30pt}\addtolength{\headsep}{30pt}}
\begin{document}
\thispagestyle{officialbibref}
\maketitle
\begin{abstract}
We introduce NameTag 3, an open-source tool and cloud-based web service for multilingual, multidataset, and multitagset named entity recognition (NER), supporting both flat and nested entities.  NameTag 3 achieves state-of-the-art results on 21 test datasets in 15 languages and remains competitive on the rest, even against larger models. It is available as a command-line tool and as a cloud-based service, enabling use without local installation. NameTag 3 web service currently provides flat NER for 17 languages, trained on 21 corpora and three NE tagsets, all powered by a single 355M-parameter fine-tuned model; and nested NER for Czech, powered by a 126M fine-tuned model. The source code is licensed under open-source MPL 2.0, while the models are distributed under non-commercial CC BY-NC-SA 4.0. Documentation is available at \url{https://ufal.mff.cuni.cz/nametag}, source code at \url{https://github.com/ufal/nametag3}, and trained models via \url{https://lindat.cz}. The REST service and the web application can be found at \url{https://lindat.mff.cuni.cz/services/nametag/}. A demonstration video is available at \url{https://www.youtube.com/watch?v=-gaGnP0IV8A}.
\end{abstract}

\section{Introduction}

Named entity recognition (NER), the task of identifying proper names such as persons, locations, and organizations in natural text, is a fundamental preprocessing step in many natural language processing (NLP) and knowledge extraction systems. While both flat and nested (embedded) NER have been extensively researched, particularly for English, many other languages still lack off-the-shelf, open-source NER tools that can be easily integrated into academic and research workflows.

We introduce NameTag 3, an open-source tool, web application, and web service for both flat and nested named entity recognition. NameTag~3 achieves state-of-the-art performance on 21 test datasets across 15 languages: Cebuano, Chinese, Croatian, Czech, Danish, English, Norwegian Bokmål, Norwegian Nynorsk, Portuguese, Russian, Serbian, Slovak, Swedish, Tagalog, and Ukrainian. Additionally, it delivers competitive results on Arabic, Dutch, German, Maghrebi, and Spanish. %

The key characteristics of NameTag 3 are:

\begin{citemize}
    \item open-source NER tool,
    \item support for both flat and nested NER,
    \item availability as command-line tool, web application, or cloud-based REST API webservice, allowing use without installation,
    \item an open-source MPL 2.0 license for code,
    \item a non-commercial CC BY-NC-SA 4.0 license for models,
    \item trained models,
    \item support for training custom models,
    \item modestly-sized models (126M or 355M),
    \item SOTA on 21 datasets in 15 languages.
\end{citemize}

Lastly, given the recent accomplishments of large language models, we also perform zero-shot and few-shot evaluations of DeepSeek-R1, demonstrating that when training data are available, NameTag 3 undoubtedly delivers substantially better performance while requiring several orders of magnitude fewer resources.

\section{Related Work}

\begin{table*}
    \centering
    \begin{tabular}{l|ccc}
    \toprule
    & NameTag 3 & Stanza & SpaCy \\
    \midrule
    Languages & 17 & 29 & 24 \\
    Architecture & fine-tuned PLM & \makecell[c]{frozen Flair embeddings,\\Bi-LSTM + CRF} & \makecell[c]{fine-tuned PLM\\or CNN}\\
    Flat NER & \YES & \YES & \YES \\
    Nested NER & \YES & \NO & \NO \\
    Single multilingual model & \YES & \NO & \YES \\
    Cross-lingual transfer & \YES & \NO & \YES \\
    Cloud-based service running & \YES & \NO & \NO \\
    \bottomrule
    \end{tabular}
    \caption{High-level technical and architectural overview of NameTag 3, Stanza, and SpaCy.}
    \label{tab:related-work-stanza}
\end{table*}

One of the most well-known NLP pipelines for NER is Stanza \cite{qi-etal-2020-stanza}, a neural-based framework developed by the Stanford NLP Group. Stanza provides pre-trained models for multiple languages.\footnote{\url{https://stanfordnlp.github.io/stanza/ner_models.html}} This pipeline is based on pre-BERT, frozen contextual character-level word embeddings \cite{akbik-etal-2018-contextual} with Bi-LSTM and CRF \cite{huang-etal-2015-bidirectional} layers on top.

Another known NLP pipeline is SpaCy \cite{spacy2}. SpaCy is a free, open-source library for advanced Natural Language Processing (NLP) in Python. SpaCy uses multitask learning with pretrained transformers like BERT in its newer models, and CNNs in its older models.

Since 2014, NameTag has provided NER for Czech and English in academic settings as NameTag 1 \cite{strakova-etal-2014-open}. In 2019, NameTag 2 \cite{strakova-etal-2019-neural} expanded to six languages --- English, German, Dutch, Spanish, Czech, and Ukrainian --- each with a separately trained model.

This publication introduces NameTag 3, which surpasses its predecessors by improving F1 scores and further expands the number of languages available. Unlike NameTag 2, which used a Bi-LSTM layer over frozen multilingual BERT embeddings, NameTag 3 fine-tunes pre-trained models with either a softmax head for flat NER or a seq2seq head for nested NER, and adds multitagset learning. 

Compared to Stanza, NameTag 3 so far supports fewer languages overall but includes some that Stanza does not cover. While Stanza employs a Bi-LSTM over frozen contextualized embeddings and trains separate models for each language, NameTag 3 takes a different approach. It is a fine-tuned PLM trained as a single joint model across multiple languages, datasets, and tagsets, enabling cross-lingual transfer even for languages not present in the training data. Additionally, NameTag 3 supports nested NER and provides a cloud-based web service.

A high-level technical and architectural overview of NameTag 3, Stanza, and SpaCy is available in Table~\ref{tab:related-work-stanza}, and the performance evaluation in F1 is presented in Table~\ref{tab:flat-results}.

\section{Data}

\begin{table*}
    \centering
    \begin{tabular}{l|l|lll}
    \toprule
                    & Flat & Nested & Nested & Nested \\
                    & Mono \& Multi & ACE 2004 & ACE 2005 & CNEC 2.0 \\
                    
    \midrule
    Encoder & XLM-R Large & RoBERTa Large & RoBERTa Large & RobeCzech Base \\
    Frozen epochs & 0 & 20 & 20 & 20 \\
    Frozen learning rate & -- & 1e-3 & 1e-3 & 1e-3 \\
    Epochs  & 30 & 60 & 50 & 20 \\
    Batch size & 8 & 8 & 16 & 4 \\
    Peak learning rate & 2e-5 & 2e-5 & 2e-5 & 2e-5 \\
    Warmup epochs & 1 & 1 & 1 & 1 \\
    Learning rate decay & cosine & cosine & cosine & cosine \\
    \bottomrule
    \end{tabular}
    \caption{Training hyperparameters.}
    \label{tab:hyperparameters}
\end{table*}

\subsection{Flat NE Datasets}
\label{sec:data-flat}

We utilized the following flat NE datasets, adhering to their official train/dev/test splits for training, tuning, and evaluation, respectively. All UNER corpora were released under the UniversalNER v1 (UNER) initiative \cite{mayhew-etal-2024-universal}.\footnote{\url{https://www.universalner.org/}} All OntoNotes 5.0 corpora follow the CoNLL-2012 train/dev/test split \cite{pradhan-etal-2012-conll} over the original OntoNotes 5.0 data.\footnote{\url{https://catalog.ldc.upenn.edu/LDC2013T19}}

\begin{citemize}
    \item \textbf{Arabic OntoNotes 5.0},
    \item \textbf{Chinese OntoNotes 5.0},
    \item \textbf{Chinese UNER GSDSIMP},
    \item \textbf{Chinese UNER GSD},
    \item \textbf{Croatian UNER SET},
    \item \textbf{Czech CNEC 2.0 CoNLL} --- In order to train and serve the Czech Named Entity Corpus 2.0 \cite{Sevcikova2007} jointly within a large multilingual model, the original annotation of the CNEC 2.0 has been harmonized to the standard 4-label tagset with \texttt{PER}, \texttt{ORG}, \texttt{LOC}, and \texttt{MISC}, resulting in an extensive simplification of the original annotation and flattening of the original nested entities.
    \item \textbf{Danish UNER DDT},
    \item \textbf{Dutch CoNLL-2002} \cite{CoNLL2002},
    \item \textbf{English OntoNotes 5.0},
    \item \textbf{English UNER EWT},
    \item \textbf{English CoNLL-2003} \cite{CoNLL2003},
    \item \textbf{German CoNLL-2003} \cite{CoNLL2003},    
    \item \textbf{Maghrebi Arabic French UNER Arabizi},
    \item \textbf{Norwegian Bokmål UNER NDT},
    \item \textbf{Norwegian Nynorsk UNER NDT},
    \item \textbf{Portuguese UNER Bosque},
    \item \textbf{Serbian UNER SET},
    \item \textbf{Slovak UNER SNK},
    \item \textbf{Spanish CoNLL-2002} \cite{CoNLL2002},
    \item \textbf{Swedish UNER Talbanken},
    \item \textbf{Ukrainian Lang-uk} --- Ukrainian Lang-uk NER corpus\footnote{\url{https://github.com/lang-uk/ner-uk}} based on the Lang-uk initiative.\footnote{\url{https://lang.org.ua/en/}} The corpus uses four classes \texttt{PER}, \texttt{ORG}, \texttt{LOC}, and \texttt{MISC}. (Please note that we harmonized the original \texttt{PERS} to a more common \texttt{PER}.)
\end{citemize}

For cross-lingual/out-of-domain evaluation on unseen languages/datasets, respectively, we used the following UNER \cite{mayhew-etal-2024-universal} test datasets: \textbf{Cebuano UNER GJA}, \textbf{Chinese UNER PUD}, \textbf{Portuguese UNER PUD}, \textbf{Russian UNER PUD}, \textbf{Swedish UNER}, \textbf{Tagalog UNER TRG}, and \textbf{Tagalog UNER Ugnayan}.

\subsection{Nested NE Datasets}

We evaluate NameTag 3 on the following nested NE corpora:

\begin{citemize}
  \item English \textbf{ACE-2004},
    \cite{Doddington}.\footnote{\url{https://catalog.ldc.upenn.edu/LDC2005T09}}
    We reuse the train/dev/test split used by most previous authors
    \cite{Lu2015,Muis2017,Wang2018}.
  \item English \textbf{ACE-2005}.\footnote{\url{https://catalog.ldc.upenn.edu/LDC2006T06}}
    Again, we use the train/dev/test split by \citet{Lu2015,Muis2017,Wang2018}.
  \item Czech \textbf{CNEC 2.0} --- Czech Named Entity Corpus 2.0 \cite{Sevcikova2007}. We use the official evaluation script distributed with the dataset, which evaluates $46$ fine-grained entity types and $4$ entity containers.
\end{citemize}

\section{Methodology}
\label{sec:methodology}

\begin{figure*}[t!]
    \centering
    \includegraphics[width=.8\hsize]{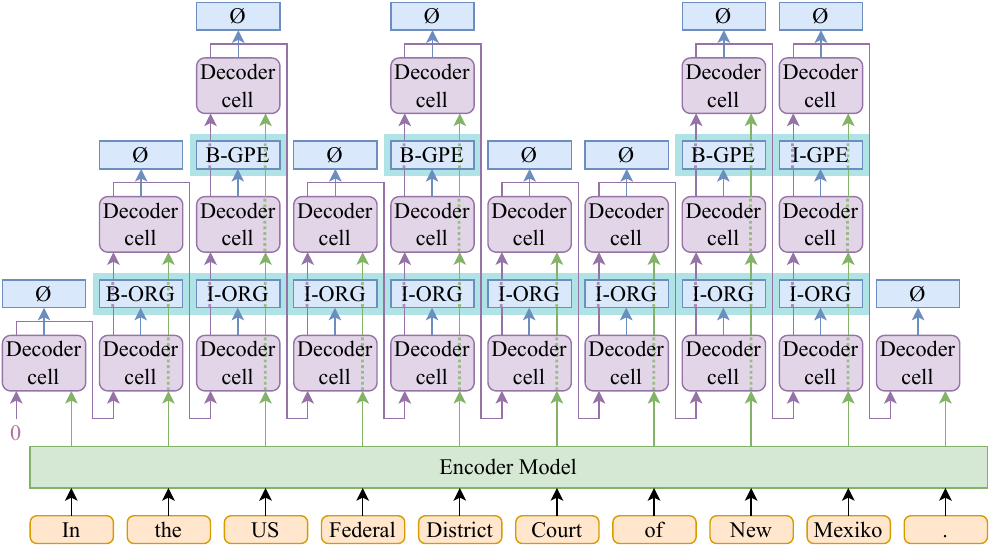}
    \caption{Visualization of the nested NER seq2seq decoder with hard attention on the current token. The example sentence is taken from ACE-2004~\citep{Doddington}.}
    \label{fig:seq2seq-decoder}
\end{figure*}

All NameTag 3 models are fine-tuned pre-trained language models of either Large (355M) or Base (126M) size. For flat NER, we apply a classification softmax head on top of the language model, while for nested NER, we use a seq2seq decoding head instead \cite{strakova-etal-2019-neural}. Both flat and nested NameTag 3 models support training on a collection of datasets, potentially in different languages. However, only NameTag 3 allows training on multiple tagsets with differing label sets.

\subsection{Flat NER}

\begin{figure}[t!]
    \centering
    \includegraphics[width=\hsize]{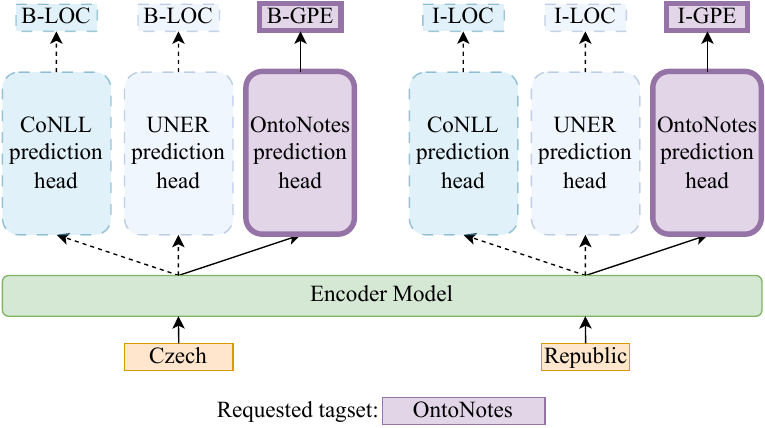}
    \caption{Visualization of the flat NER classification heads for multiple tagsets.}
    \label{fig:multitagset}
\end{figure}

For flat NER, NameTag 3 enables multitagset learning by assigning a separate classification head to each tagset and jointly training the encoder and all classification heads. During inference, the classification head corresponding to the requested tagset is used, ensuring that only valid tags are predicted, see visualization in Fig.~\ref{fig:multitagset}.

The currently supported tagsets are:

\begin{citemize}
\item \texttt{conll}: The CoNLL-2002 and CoNLL-2003 \cite{CoNLL2002,CoNLL2003} tagset,
\item \texttt{uner}: The Universal NER v1 \cite{mayhew-etal-2024-universal} tagset,
\item \texttt{onto}: The OntoNotes 5.0 tagset.
\end{citemize}

The NameTag 3 multilingual flat NER model was trained on the training portions of the flat NER datasets described in Sec.~\ref{sec:data-flat}. Training batches were constructed using square root temperature sampling, in which the examples from the corpora are sampled into training batches proportionally to the square root of the number
of their sentences, similarly to \citet{van-der-goot-etal-2021-massive}.
This approach effectively downsamples the largest corpora while upsampling the smallest ones. To achieve balanced performance across all datasets, we use a macro span-based F1 score with uniform weighting as our evaluation objective. The training hyperparameters are described in Table~\ref{tab:hyperparameters}.

\subsection{Nested NER}

For nested named entity recognition, we replace the flat softmax classification head with a sequence-to-sequence (seq2seq) decoder head \cite{strakova-etal-2019-neural}, see visualization in Figure~\ref{fig:seq2seq-decoder}. This decoder generates a sequence of linearized (flattened) nested output labels for each input token embedded by the pre-trained LM encoder. The Transformer encoder and seq2seq decoder weights are fine-tuned jointly. Before fine-tuning, we perform a few pre-training epochs with frozen Transformer encoder weights to allow the seq2seq decoder to adjust to them. This helps ensure a smoother transition into fine-tuning. The training hyperparameters are described in Table~\ref{tab:hyperparameters}.

\section{Results}

\begin{table*}[t]
    \centering
    \setlength{\tabcolsep}{3.8pt}
    \begin{tabular}{l|cc|cc|clr}
    \toprule
    & Mono & Multi & Stanza & SpaCy & SOTA & SOTA & \makecell[c]{SOTA}\\
    Corpus & F1 & F1 & F1 & F1 & F1 & Ref. & \makecell[c]{Params}\\
    \midrule
Arabic OntoNotes v5    & 75.50 & 74.20	& --- & --- & \textbf{76.40} & \citet{aloraini-etal-2020-neural} & 136M \\
Chinese OntoNotes v5   & \textbf{81.76} & 81.63	& 79.2\rlap{$^\heartsuit$} & --- &  80.20 & \citet{li2023named} & 147M \\
Chinese UNER GSDSIMP   & 88.99 & \textbf{90.99}	& --- & --- & 89.40 & \citet{mayhew-etal-2024-universal}\textsuperscript{\ddag} & 355M \\
Chinese UNER GSD       & 90.14 & \textbf{91.53}	& --- & --- & 89.50 & \citet{mayhew-etal-2024-universal}\textsuperscript{\ddag} & 355M \\
Croatian UNER SET      & 94.08 & \textbf{95.55}	& --- & --- & 95.00 & \citet{mayhew-etal-2024-universal}\textsuperscript{\ddag} & 355M \\
Czech CNEC 2.0 CoNLL   & 85.31 & \textbf{86.24}	& --- & --- & --- & --- & --- \\
Danish UNER DDT        & 87.21 & \textbf{89.75}	& --- & --- & 88.10 & \citet{mayhew-etal-2024-universal}\textsuperscript{\ddag} & 355M \\
Dutch CoNLL-2002	   & 95.16 & 94.93	& 89.2\rlap{$^\heartsuit$} & --- & \textbf{95.70} & \citet{wang-etal-2021-automated} & 1117M\rlap{\textsuperscript{\dag}} \\
English OntoNotes v5   & 90.22 & 90.19	& 88.8\rlap{$^\heartsuit$} & 89.8\rlap{$^\diamondsuit$} & \textbf{92.07} & \citet{li-etal-2020-dice} & 336M \\
English UNER EWT	   & 86.27 & \textbf{87.03}	& --- & --- & 85.80 & \citet{mayhew-etal-2024-universal}\textsuperscript{\ddag} & 355M \\
English CoNLL-2003     & 93.80 & 94.09	& 92.1\rlap{$^\heartsuit$} & 91.6\rlap{$^\diamondsuit$} & \textbf{94.60} & \citet{wang-etal-2021-automated} & 1853M\rlap{\textsuperscript{\dag}} \\
German CoNLL-2003	   & 87.77 & 87.48	& 81.9\rlap{$^\heartsuit$} & --- & \textbf{88.38} & \citet{wang-etal-2021-automated} & 1108M\rlap{\textsuperscript{\dag}} \\
Maghrebi UNER Arabizi  & 72.77 & 84.49	& --- & --- & \textbf{86.20} & \citet{mayhew-etal-2024-universal}\textsuperscript{\ddag} & 355M \\
Norw. Bokmål UNER NDT  & 93.97 & \textbf{95.83}	& --- & --- & --- & --- & --- \\
Norw. Nynorsk UNER NDT & 93.71 & \textbf{94.51}	& --- & --- & --- & --- & --- \\
Portuguese UNER Bosque & \textbf{91.18} & 90.89	& --- & --- & 90.40 & \citet{mayhew-etal-2024-universal}\textsuperscript{\ddag} & 355M  \\
Serbian UNER SET	   & 94.85 & \textbf{97.10}	& --- & --- & 96.60 & \citet{mayhew-etal-2024-universal}\textsuperscript{\ddag} & 355M \\
Slovak UNER SNK	       & 86.79 & \textbf{88.46} & --- & --- & 85.50 & \citet{mayhew-etal-2024-universal}\textsuperscript{\ddag} & 355M \\
Spanish CoNLL-2002	   & 88.95 & 90.29	& 88.1\rlap{$^\heartsuit$} & --- & \textbf{90.40} & \citet{wang-etal-2021-automated} & 1105M\rlap{\textsuperscript{\dag}} \\
Swedish UNER Talbanken & 90.73 & \textbf{91.79}	& --- & --- & 88.30 & \citet{mayhew-etal-2024-universal}\textsuperscript{\ddag} & 355M \\
Ukrainian Lang-uk	   & 90.45 & \textbf{92.88}	& 86.1\rlap{$^\heartsuit$} & --- & 88.73 & NameTag 2 & 110M \\
    \bottomrule
    \end{tabular}
    \caption{NameTag 3 flat NER span-based micro F1 with the monolingual (Mono) models and the multilingual (Multi) model of 355M params. We report the highest F1 scores from the respective leaderboards on \url{https://paperswithcode.com/} where available. \dag \citet{wang-etal-2021-automated} use a concatenation of multiple embeddings, incl. several Base and Large. \ddag For \citet{mayhew-etal-2024-universal}, we report the better result from the ``in-language'' (Table 4) and ``all'' (Table 5). $\heartsuit$ \url{https://stanfordnlp.github.io/stanza/ner_models.html}. $\diamondsuit$ \url{https://spacy.io/usage/facts-figures}.}
    \label{tab:flat-results}
\end{table*}

\begin{table*}[t!]
    \centering
    \begin{tabular}{lc}
    \toprule
    Model & F1 \\
    \midrule
    ChatGPT 3.5 zero-shot \cite{xie-etal-2024-self} & 68.97\textsuperscript{\dag} \\
    ChatGPT 3.5 ICL with self-annotated demonstrations \cite{xie-etal-2024-self} & 74.99\textsuperscript{\dag} \\
    \midrule
    DeepSeek R1 32B zero-shot & 64.33 \\
    DeepSeek R1 32B 5-shot & 74.26 \\
    \midrule
    DeepSeek R1 70B zero-shot & 67.97\\
    DeepSeek R1 70B 5-shot & 74.00\\
    \midrule
    NameTag 3 & \textbf{94.09} \\
    \bottomrule
    \end{tabular}
    \caption{Comparison of NameTag 3 with NER performed by prompting LLMs on the (entire) English CoNLL-2003 test dataset (3\,684 sentences). \dag \citet{xie-etal-2024-self} report the mean of two samples of 300 sentences.}
    \label{tab:llm-results}
\end{table*}

\begin{table*}[t]
    \centering
    \begin{tabular}{llccc}
    \toprule
    Model & GPU & Batch & Sentences per sec. & Time \\
    \midrule
    DeepSeek R1 70B zero-shot & AMD MI210 & 1 & 0.05 & 23h \\
    DeepSeek R1 70B 5-shot & AMD MI210 & 1 & 0.04 & 25h \\
    \midrule
    DeepSeek R1 32B zero-shot & AMD MI210 & 1 & 0.08 & 13h \\
    DeepSeek R1 32B 5-shot & AMD MI210 & 1 & 0.06 & 16h \\
    \midrule
    NameTag 3 & AMD MI210 & 1 & 801 & 4.6s \\
    NameTag 3 & AMD MI210 & 8 & 784 & 4.7s \\
    \midrule
    NameTag 3 & NVIDIA A30 & 1 & 646 & 5.7s \\
    NameTag 3 & NVIDIA A30 & 8 & 801 & 4.6s \\
    \bottomrule
    \end{tabular}
    \caption{Sentence throughput in sentences per second of the NameTag 3 REST API and Deep Seek REST API by predicting the (entire) English CoNLL-2003 test dataset (3\,684 sentences).}
    \label{tab:throughput}
 \end{table*}

\begin{table*}[t]
    \centering
    \begin{tabular}{l|c|clll}
    \toprule
           &    & SOTA & SOTA & SOTA \\  
    Corpus & F1 & F1   & Ref. & Params. \\
    \midrule
    ACE-2004 & 88.39 & \textbf{88.72} & \citet{shen-etal-2023-promptner} & 345M \\

    ACE-2005 & 87.21 & \textbf{88.83}          & \citet{yuan-etal-2022-fusing} & 223M \\

    CNEC 2.0 & \textbf{86.39} & 83.44 & NameTag 2 & 110M \\
    \bottomrule
    \end{tabular}
    \caption{NameTag 3 nested NER span-based micro F1. CNEC 2.0 is the only corpus modeled with a Base-sized monolingual Czech encoder RobeCzech Base (126M). The ACE models are based on RoBERTa Large (355M).}
    \label{tab:nested-results}
\end{table*}

\begin{table}[t]
    \centering
    \begin{tabular}{l|c|c}
    \toprule
    Corpus & F1 & SOTA F1 \\
    \midrule
    Cebuano UNER GJA & \textbf{96.97} & 82.2 \\
    Chinese UNER PUD & \textbf{89.35} & 86.0 \\
    Portuguese UNER PUD & \textbf{91.77} & 87.5 \\
    Russian UNER PUD & \textbf{75.51} & 73.6 \\
    Swedish UNER PUD & \textbf{91.27} & 88.0 \\
    Tagalog UNER TRG  & \textbf{97.78} & 83.7 \\
    Tagalog UNER Ugnayan & 75.00 & \textbf{76.1} \\
    \bottomrule
    \end{tabular}
    \caption{Cross-lingual/out-of-domain evaluation on unseen languages/datasets predicted by cross-lingual transfer with the NameTag 3 multilingual flat model of 355M parameters. The metric is flat NER span-based micro F1. Previous SOTA F1 are from \citet{mayhew-etal-2024-universal}, whose multilingual model is also of 355M.}    
    \label{tab:ood-results}
\end{table}

\subsection{Flat NER}

Table~\ref{tab:flat-results} presents NameTag 3 span-based micro F1 with the monolingual (Mono) models and the multilingual (Multi) model of 355M params.

Alongside our results, we report the highest F1 scores from the respective leaderboards on \url{https://paperswithcode.com/} where available, and/or the current state-of-the-art academic baselines; many of these models originate from academic research and do not provide ready-to-use tools, and/or often rely on significantly larger model capacities in terms of parameter count.

Apart from the state-of-the-art models, we also compare NameTag 3 to popular NLP toolkits supporting named entity recognition: Stanza~\cite{qi-etal-2020-stanza} and SpaCy~\cite{spacy2}. Our system surpasses both these toolkits on all the datasets where pretrained models are available.\footnote{Both Stanza and SpaCy provide models for more languages, but trained on different datasets with possibly different tag sets, preventing direct comparison on more languages.}

Table~\ref{tab:ood-results} presents out-of-domain evaluation on unseen languages/datasets by cross-lingual transfer. The accompanying previous SOTA results are from \citet{mayhew-etal-2024-universal}.

\textbf{LLM Evaluation}~~~We include comparison of NameTag 3 with LLMs in Table~\ref{tab:llm-results} to demonstrate that fine-tuning ``smaller'' models (355M vs. 70B parameters) is still worthwhile even in the era of generative AI. We prompt DeepSeek-R1 70B \cite{deepseekai2025deepseekr1incentivizingreasoningcapability}, currently one of the best available open-source sub-100B LLMs,\footnote{Our goal was to evaluate the best available replicable model that can run without enormous resources in order to be a viable NER system alternative.} in zero-shot and 5-shot settings, and we also reprint similar prompting experiments on ChatGPT 3.5 reported in literature \cite{xie-etal-2024-self}. NameTag 3, a fine-tuned 355M model, achieves 20 percent points higher F1 score while being more than 10,000 times faster, as demonstrated in performance measurements Tab~\ref{tab:throughput}. Therefore, when training data are available, NameTag 3 constitutes a much more accessible and practical system, allowing users to keep processed data private using only a single consumer-grade GPU. The complete script for LLM evaluation including the used prompts and few-shot example selection is available at \url{https://github.com/ufal/nametag3/tree/acl2025/llm_baseline}.

\subsection{Nested NER}

\looseness-1
Table~\ref{tab:nested-results} shows the NameTag 3 nested NER results, evaluated as span-based micro F1. NameTag 3 with the seq2seq head for nested NER achieves state-of-the-art results on the canonical Czech nested corpus with 46 entity types and 4 containers, while reaching near-SOTA results for English nested corpora.

\section{Conclusions}

We introduced NameTag 3, a multilingual, open-source named entity recognition tool for both flat and nested NER. It is available as a command-line tool (\url{https://github.com/ufal/nametag3}) and as a web application with a cloud-based REST API (\url{https://lindat.mff.cuni.cz/services/nametag}). NameTag 3 includes pre-trained models and supports custom training.

NameTag 3 demonstrates state-of-the-art performance on 21 test datasets across 15 languages: Cebuano, Chinese, Croatian, Czech, Danish, English, Norwegian Bokmål, Norwegian Nynorsk, Portuguese, Russian, Serbian, Slovak, Swedish, Tagalog, and Ukrainian, while also performing well in Arabic, Dutch, German, Maghrebi, and Spanish.

The tool is released under the open-source MPL 2.0 license, with models distributed under non-commercial CC BY-NC-SA 4.0.

We hope NameTag 3 will be particularly valuable for the academic community and researchers working with multilingual NLP and non-English texts.

\section*{Limitations}

Since NameTag 3 classifies into a predefined set of named entity classes, it is not susceptible to issues generally associated with generative AI, such as hallucinations or the production of misleading or harmful information.

By jointly training on 21 datasets across 17 languages, NameTag 3 is less prone to biases that typically affect monolingual or culturally homogeneous models. We hope that this multilingual approach helps mitigate issues like overrepresentation of Western-centric names and gender imbalances in named entity distributions.

However, most of our training datasets are written in Latin scripts, with the exception of Chinese (three datasets), Arabic (two datasets), and Ukrainian (one dataset). We recognize the need to further improve coverage by incorporating additional languages.

This brings us to an important limitation: As a supervised, fine-tuned model, NameTag 3 relies on gold-standard, manually annotated training data. Expanding the diversity and volume of such data is crucial for further improving performance across languages and domains.

In future work, we plan to expand our set of manually annotated training data while also exploring silver-standard, semi-automated data to further increase the volume of training material.

\section*{Acknowledgments}

This research was supported by the Johannes Amos Comenius Programme (P JAC) project No. CZ.02.01.01/00/22\_008/0004605, Natural and anthropogenic georisks,  and it has also been supported by the Ministry of Education, Youth and Sports of the Czech Republic, Project No. LM2023062 LINDAT/CLARIAH-CZ. The work described herein uses resources hosted by the LINDAT/CLARIAH-CZ Research Infrastructure (projects LM2018101 and LM2023062, supported by the Ministry of Education, Youth and Sports of the Czech Republic).

\bibliography{custom}

\end{document}